\documentclass[10pt,twocolumn,letterpaper]{article}

\usepackage[pagenumbers]{iccv} 

\definecolor{iccvblue}{rgb}{0.21,0.49,0.74}
\usepackage[pagebackref,breaklinks,colorlinks,allcolors=iccvblue]{hyperref}

\def\paperID{5767} 
\def\confName{ICCV}
\def\confYear{2025}

\title{Harnessing Caption Detailness for Data-Efficient Text-to-Image Generation}

\author{Xinran Wang$^{1\star}$\quad Muxi Diao$^{1,2\star}$ \quad Yuanzhi Liu$^{1}$ \quad Chunyu Wang$^{\xi}$ \\ Kongming Liang$^{1\dag}$ \quad Zhanyu Ma$^{1}$ \quad Jun Guo$^{1}$ \\ 
$^{1}$Beijing University of Posts and Telecommunications \quad $^{2}$Zhongguancun Academy\\
{\tt\small \{wangxr, dmx, liuyuanzhi, liangkongming, mazhanyu, guojun\}@bupt.edu.cn} \\ {\tt \small chunyu.wangdlut@gmail.com} 
}

\begin{document}

\twocolumn[{%
\renewcommand\twocolumn[1][]{#1}%
\maketitle


\vspace{-3.2em}
\begin{center}
    \centering
    \captionsetup{type=figure}
    \includegraphics[width=\textwidth]{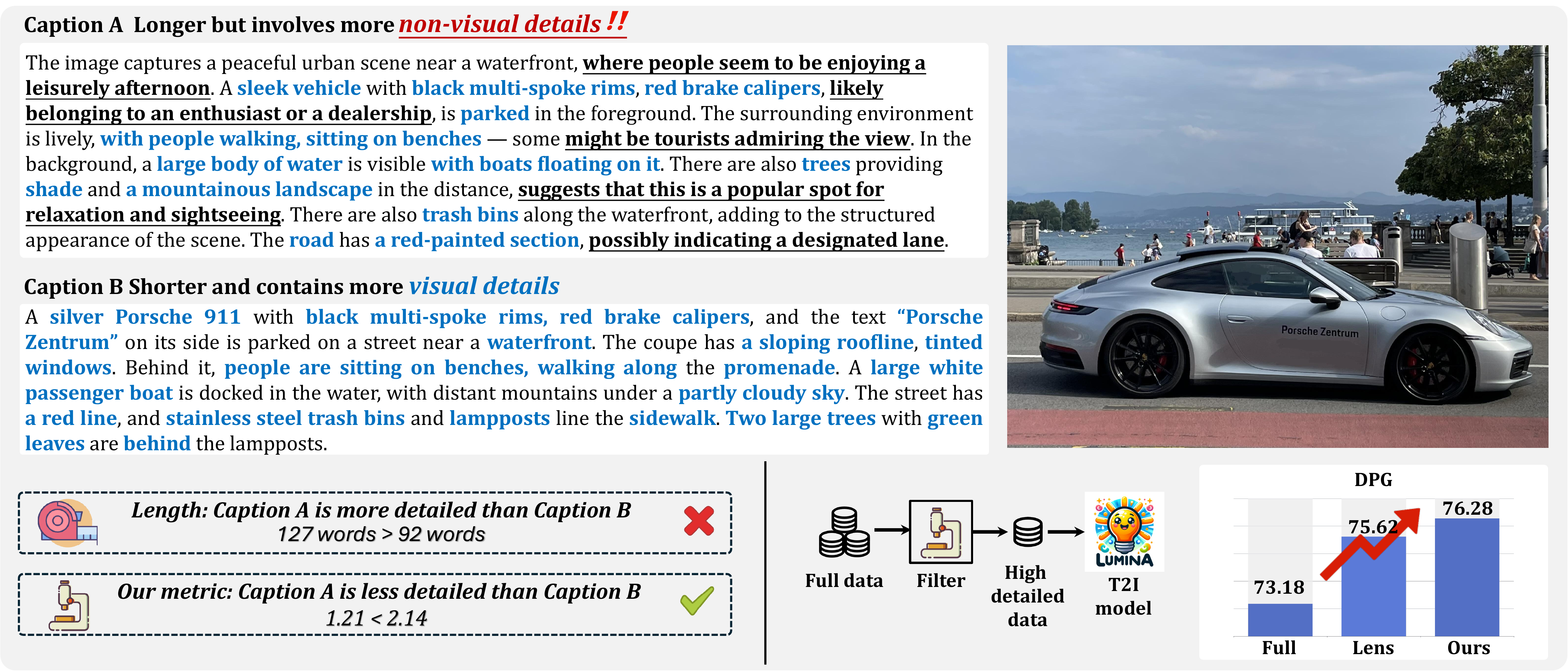}
    \vspace{-2em}
    \captionof{figure}{Caption length is not a perfect indicator of caption detailness. In this paper, we propose a new metric to more effectively estimate caption detailness. Using this metric to select T2I training data surpasses both full-data training and length-based selection methods on the dense prompt graph (DPG) benchmark.}
    \label{fig:insight}
\end{center}%
}]

\maketitle

\begin{abstract}

\vspace{-2.3em}

Training text-to-image (T2I) models with detailed captions can significantly improve their generation quality. Existing methods often rely on simplistic metrics like caption length to represent the detailness of the caption in the T2I training set. In this paper, we propose a new metric to estimate caption detailness based on two aspects: image coverage rate (ICR), which evaluates whether the caption covers all regions/objects in the image, and average object detailness (AOD), which quantifies the detailness of each object's description. Through experiments on the COCO dataset using ShareGPT4V captions, we demonstrate that T2I models trained on high-ICR and -AOD captions achieve superior performance on DPG and other benchmarks. Notably, our metric enables more effective data selection—training on only 20\% of full data surpasses both full-dataset training and length-based selection method, improving alignment and reconstruction ability. These findings highlight the critical role of detail-aware metrics over length-based heuristics in caption selection for T2I tasks.
\end{abstract}

\begingroup
\renewcommand\thefootnote{}\footnotetext{$\star$ Equal contribution}
\renewcommand\thefootnote{}\footnotetext{$\xi$ Independent researcher}
\renewcommand\thefootnote{}\footnotetext{$\dag$ Corresponding author}
\endgroup

\section{Introduction}


Recent text-to-image (T2I) generation advancements have showcased the remarkable potential of converting textual descriptions into vivid images. The training paradigm for T2I generation models typically involves encoding image captions through text encoders to establish conditional control over generated content. Although current T2I models \cite{rombach_high-resolution_2022, podell_sdxl:_2024, reed_generative_2016, ramesh_zero-shot_2021, chen_pixart-textbackslashalpha:_2024} achieve impressive results, they rely on massive simplistic caption datasets \cite{schuhmann_laion-5b:_2022, sharma_conceptual_2018, fleet_microsoft_2014, young_image_2014} in their training stage. This reliance introduces a data efficiency problem—models trained on large-scale datasets still struggle with generating complex visual scenes by inputting long, richly detailed textual prompts. To solve this problem, many works apply more detailed image captions \cite{li_densefusion-1m:_2024, pi_image_2024, bonilla-salvador_pixlore:_2024, garg_imageinwords:_2024, onoe_docci:_2025, urbanek_picture_2024} to enhance the ability of T2I models on long prompts following. However, manually annotating images with detailed captions is prohibitively expensive. To reduce costs, many studies \cite{betker_improving_2023, liu_playground_2024, li_what_2024, wang_emu3:_2024, liu_improving_2025} leverage multimodal large language models (MLLMs) to generate detailed image captions beyond traditional ones. These models not only describe primary objects but also capture attributes, spatial relationships, and contextual interactions, providing stronger control signals for training T2I models.

However, measuring caption detailness remains a challenge. Previous works \cite{li_what_2024, lai_revisit_2025} simply represent detailness through caption length, assuming longer captions inherently contain richer information. As shown in \Cref{fig:insight}, length is not the perfect proxy for caption detailness since it suffers from two limitations: 1) Long descriptions may also miss certain areas of the image, and 2) Verbose captions may contain redundant details that are irrelevant to the visual information in the image. To address this gap, we propose a new caption detailness metric considering two aspects: \textbf{image coverage rate (ICR)} and \textbf{average object detailness (AOD)}. ICR quantifies the proportion of regions / objects in the image accurately described in the caption, while AOD assesses the granularity of object descriptions by considering both attributes and relations. 


To analyze the influence of these two aspects, we first conducted an experiment to investigate the relationship between AOD, ICR, and T2I performance. Using the detailed ShareGPT4V \cite{leonardis_sharegpt4v:_2025} captions for COCO \cite{fleet_microsoft_2014}, we generated multiple sets of captions with varying ICR and AOD proportions. Then we trained the same T2I model on each set and evaluated their performance. The results revealed a clear positive correlation between model performance and the two metrics.

To further validate the practical utility of the proposed metric, we employed it for data filtering on the ShareGPT4V-COCO dataset. Notably, training on no more than $20\%$ of the full data selected by our metrics resulted in higher image-text consistency and better image reconstruction effects compared to training on the entire dataset, highlighting their effectiveness for selecting high-quality training data. Moreover, our approach outperformed length-based filtering strategies, underscoring that \textbf{caption length alone is not a reliable indicator of caption detailness}.

Our contributions can be summarized as follows: 

(1) \textbf{Novel caption detailness metrics}: We propose a new metric which considers two aspects: ICR quantifies the proportion of objects accurately described in the caption, while AOD assesses the granularity of object descriptions by considering both attributes and relationships. 

(2) \textbf{Uncover the positive correlation trend}: We design experiments to show that higher AOD and ICR values in training captions lead to improved text-to-image generation.

(3) \textbf{Caption detailness-based data filtering strategy}: Utilizing AOD and ICR for training data selection surpasses both full-data and length-based filtering methods, enhancing image-text consistency and improving image reconstruction performance.

\section{Related Work}

\subsection{Detailed Image Captioning}

Recent image captioning models leverage large language models (LLMs) \cite{brown_language_2020, raffel_exploring_2020, chiang_vicuna:_2023, aimeta_llama_2024, glm_chatglm:_2024,  team_gemma:_2024, deepseek-ai_deepseek-r1:_2025}, which excel at next token prediction, as the decoder and the vision transformers (ViTs) \cite{dosovitskiy_image_2020, radford_learning_2021} as the visual encoder to process image inputs. The integration of these components forms multimodal large language models (MLLMs) \cite{dai_instructblip:_2023, liu_visual_2023, bai_qwen-vl:_2023, wang_qwen2-vl:_2024, zhang_omg-llava:_2024, leonardis_sharegpt4v:_2025}. After training on synthetic datasets, MLLMs demonstrate enhanced capabilities to produce long-form, detailed image descriptions that surpass conventional captioning outputs. Existing evaluation metrics for image captions predominantly focus on linguistic quality—assessing aspects such as text-image consistency, grammatical correctness, fluency, and syntactic structure—using measures like BLEU \cite{papineni_bleu:_2002}, ROUGE \cite{lin_rouge:_2004}, METEOR \cite{banerjee_meteor:_2005}, CIDEr \cite{vedantam_cider:_2015}, CLIP Score \cite{hessel_clipscore:_2021}, CAPTURE Score \cite{dong_benchmarking_2024} and other caption evaluation metrics \cite{jing_faithscore:_2024, lu_benchmarking_2024}. While these metrics are effective for evaluating text coherence and language quality, they often overlook an essential factor: the density of visual information contained within the caption. This oversight means that even captions with high linguistic scores may fail to capture the rich visual details necessary for guiding downstream tasks, such as text-to-image generation. Recent research efforts have begun to address this limitation by integrating question answering and visual grounding into evaluation frameworks \cite{jing_faithscore:_2024, pi_image_2024, cho_davidsonian_2024, lu_benchmarking_2024}. However, a systematic metric that quantifies the level of visual detail remains underdeveloped. In this paper, we introduce two metrics for estimating the caption detailness.

\subsection{Text-to-image Generation}

Text-to-image generation has rapidly evolved, driven by breakthroughs in deep learning architectures, including generative adversarial networks (GANs) \cite{goodfellow_generative_2014, heusel_gans_2017, goodfellow_generative_2020} and more recently, diffusion models \cite{ho_denoising_2020, song_denoising_2021, ho_classifier-free_2021, rombach_high-resolution_2022, peebles_scalable_2023, esser_scaling_2024, xie_boxdiff:_2023, ding_cogview2:_2022, zhang_iti-gen:_2023, liu_unsupervised_2023}. Early approaches laid the groundwork by conditioning image synthesis on class categories or short textual prompts, yet they often produced images that were limited in detail and fidelity. Subsequent models \cite{li_hunyuan-dit:_2024, gao_lumina-t2x:_2025, zhuo_lumina-next_2024}, particularly those incorporating diffusion transformers \cite{peebles_scalable_2023} and large language models as text encoders with long text encoding ability, have achieved higher resolution and greater semantic consistency. Despite these advancements, a persistent challenge remains: generating images from long, detailed textual descriptions. Many current models perform well with short, simplistic prompts but struggle when confronted with more elaborate and nuanced text. In this paper, our study aims to use the caption detailness metric to filter small but highly detailed descriptions to improve the text-to-image model's ability on long prompt following.

\begin{figure*}[ht]
    \centering
    \includegraphics[width=\linewidth]{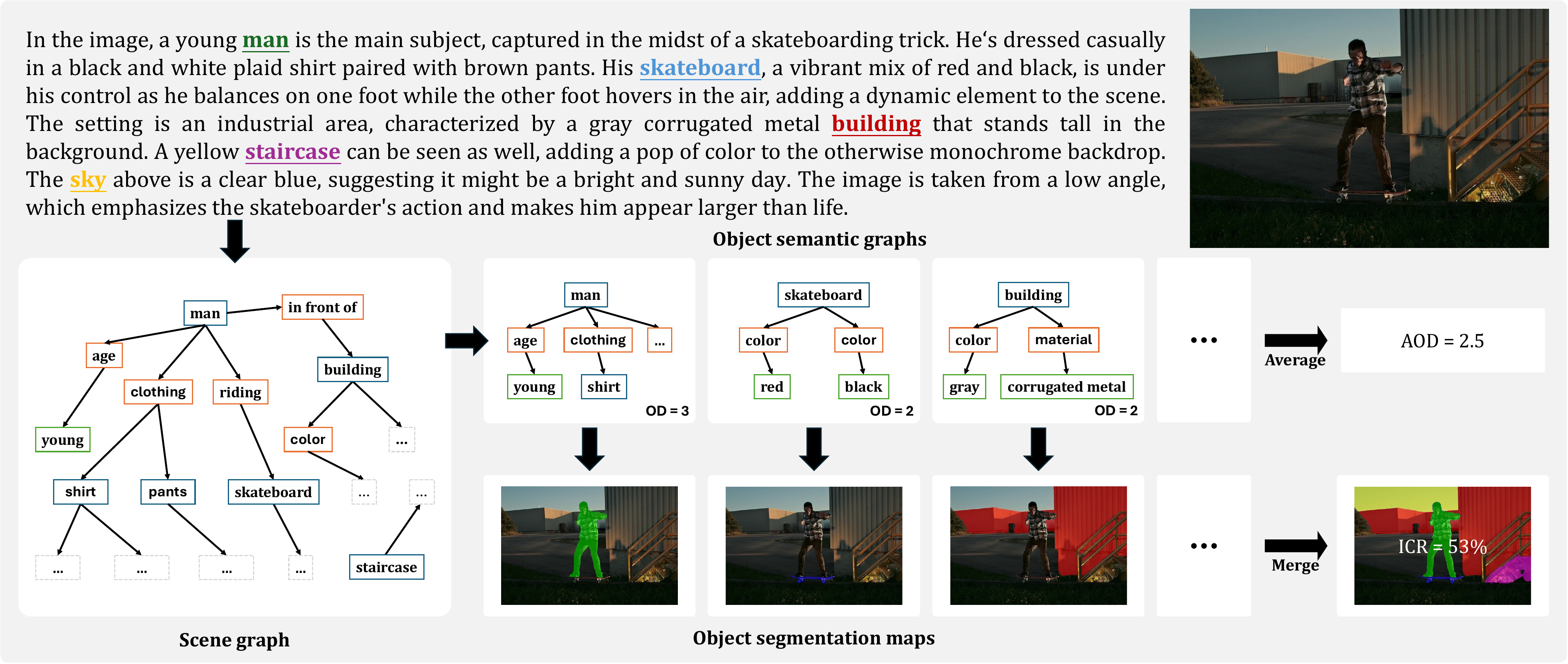}
    \caption{Illustration of calculating average object detailness (AOD) and image coverage rate (ICR). The image caption is first parsed into a scene graph, from which we extract the semantic graph for each object. A segmentation model provides object masks. AOD is computed as the average number of triplets per object graph, while ICR is determined by the total area ratio of all objects.}
    \label{fig:intro}
\end{figure*}

\section{Caption Detailness Estimation}

In this section, we first point out the inherent drawbacks of using length to represent caption detailness. Then we precisely show how to accurately calculate the image coverage rate and average object detailness based on the caption's scene graph. Finally, we thoroughly introduce the caption detailness, which integrates the above two metrics while penalizing verbosity or irrelevant content. The whole process of our caption estimation method is shown in \Cref{fig:intro}.

\subsection{Problem Formulation}

Given a training dataset of image-caption pairs $D=\{x_i, c_i\}_{i=1}^N$, we aim to estimate the caption detailness in training data. Therefore, we must first clearly define ``detailness'' - a concept often conflated with caption length in prior work. However, textual length alone is not a perfect proxy for detailness. Consider two failure modes: (1) Captions containing redundant or image-irrelevant content (e.g., ``\textcolor{black}{\textit{A man in a red jacket is walking towards a restaurant down the street.}} \textcolor{blue}{\textit{The person is probably hungry.}},'' where the \textcolor{blue}{blue-colored} speculative clause adds no visual information), or (2) Verbose phrasing that inflates token count without enhancing visual specificity, such as repetitively stating ``\textit{There is a \textcolor{blue}{very big, enormous, gigantic} skyscraper. Cleaners are cleaning the glass on the \textcolor{blue}{very big, enormous, gigantic} skyscraper.}'' instead of concisely describing ``\textit{Cleaners are cleaning the glass on an \textcolor{blue}{enormous} skyscraper.}''.

Theoretically, an ideal detailed caption should achieve two goals: (1) Fully enumerate all objects present in the image, and (2) Characterize each object through its attribute dimensions (e.g., color, material, pose) and relational contexts (e.g., spatial arrangements, interactions). This leads us to formalize caption detailness through dual perspectives: image coverage rate (the proportion of detectable objects/scenes described) and average object detailness (the average number of attributes and relations per object).

\subsection{Scene Graph Parsing}
To estimate these two metrics, we first need to decompose the caption into concept level. The scene graph \cite{xu_scene_2017, kim_llm4sgg:_2024, li_pixels_2024} is a good semantic representation composed of concepts. Inspired by \cite{dong_benchmarking_2024, cho_davidsonian_2024, leibe_spice:_2016, li_factual:_2023, jing_faithscore:_2024, lu_benchmarking_2024, wang_detailed_2025}, we apply a scene graph parser to transform the unstructured description into a structured semantic graph. 
Given any caption $c$, we can parse it into a scene graph: 

\begin{equation}
    G(c)=\langle O(c), R(c), A(c)\rangle    
\end{equation}

\noindent where $O(c)$ is the set of object mentions in $c$, $R(c)$ is the set of relations between objects and $A(c)$ is the set of object attributes. Here we apply Llama3-70B \cite{aimeta_llama_2024} with design instruction as the scene graph parser, the instruction can be seen in the supplementary material. In the next two sections, we will respectively introduce how to evaluate the image coverage rate and average object detailness of a caption based on the semantic graph.

\subsection{Image Coverage Rate}

The image coverage rate (ICR) serves as a crucial metric to assess the comprehensiveness of a generated caption in describing objects/regions in an image. A higher ICR suggests that the caption captures a larger proportion of objects, while a lower ICR may indicate missing or ungrounded objects. A simple way to calculate the ICR is to use a segmentation model to ground all caption-mentioned objects. Formally, given an image $x$, and the object set $O(c)$ mentioned by its caption $c$, we apply a segmentation model, e.g., LISA \cite{lai_lisa:_2024} to generate a segmentation mask for each object in $O(c)$. Then the image coverage rate of image $x$ can be calculated by the union area ratio of all object masks to the area of the whole image: 

\begin{equation}
    \text{ICR}(x, c) = \frac{\text{Area}(\bigcup_{j=1}^{|O(c)|}\left\{s_j\right\})}{\text{Area}(x_i)},
\end{equation}

\noindent where $s_j$ is the segmentation mask of the object $j$ and $\text{Area}(\cdot)$ is the area calculation function.

\subsection{Average Object Detailness}

The average object detailness (AOD) reflects the level of detail of each object in the image being described. In a semantic graph, the number of edges connected to an object node indicates its detail level. Specifically, edges can represent either object attributes or relationships between objects. Given an object $o_j \in O(c)$ in the image $x$, we define its relation and attribute count as follows:
\begin{equation}
\begin{aligned}
     \mathcal{D}_R(o_j, c) &= \left|\left\{\left(o_j, r, o_k\right) \mid o_k \in O(c), r \in R(c)\right\}\right|, \\
    \mathcal{D}_A(o_j, c) &=  |\{(o_j, a) \mid a \in A(c)\}|.
\end{aligned}
\end{equation}

\noindent The total detailness score for the image $x$ is then computed as the average across all objects:
\begin{equation}
    \text{AOD}(x, c) = \frac{\sum_{j=1}^{|O(c)|}(D_R(o_j, c) + D_A(o_j, c))}{|O(c)|}.
\end{equation}

\noindent Specifically, a higher AOD score indicates that objects in an image are described with more attributes and relationships, which means the associated caption captures finer details of the visual scene.

\subsection{Caption Detailness}

To account for the differing value ranges of ICR and AOD, we combine them into a unified metric for caption detailness while penalizing verbosity or non-visual content:

\begin{equation}
\label{eq:detailness}
    \text{CD}(x, c) = \frac{\text{ICR}(x, c) \times \text{AOD}(x, c)}{\text{Length(c)}},
\end{equation}
\noindent where is $\text{Length}(\cdot)$ is the word counting function.
This formulation ensures that longer, potentially verbose captions with extraneous content receive a lower score, emphasizing concise and visually relevant descriptions.


\section{Experiments}

In this section, we first conduct experiments to assess the impact of the proposed ICR and AOD on the performance of text-to-image generation. Then we conduct a practical experiment to demonstrate how these metrics help identify and select detailed captions.








\begin{figure}[t]
    \centering
    \includegraphics[width=\linewidth]{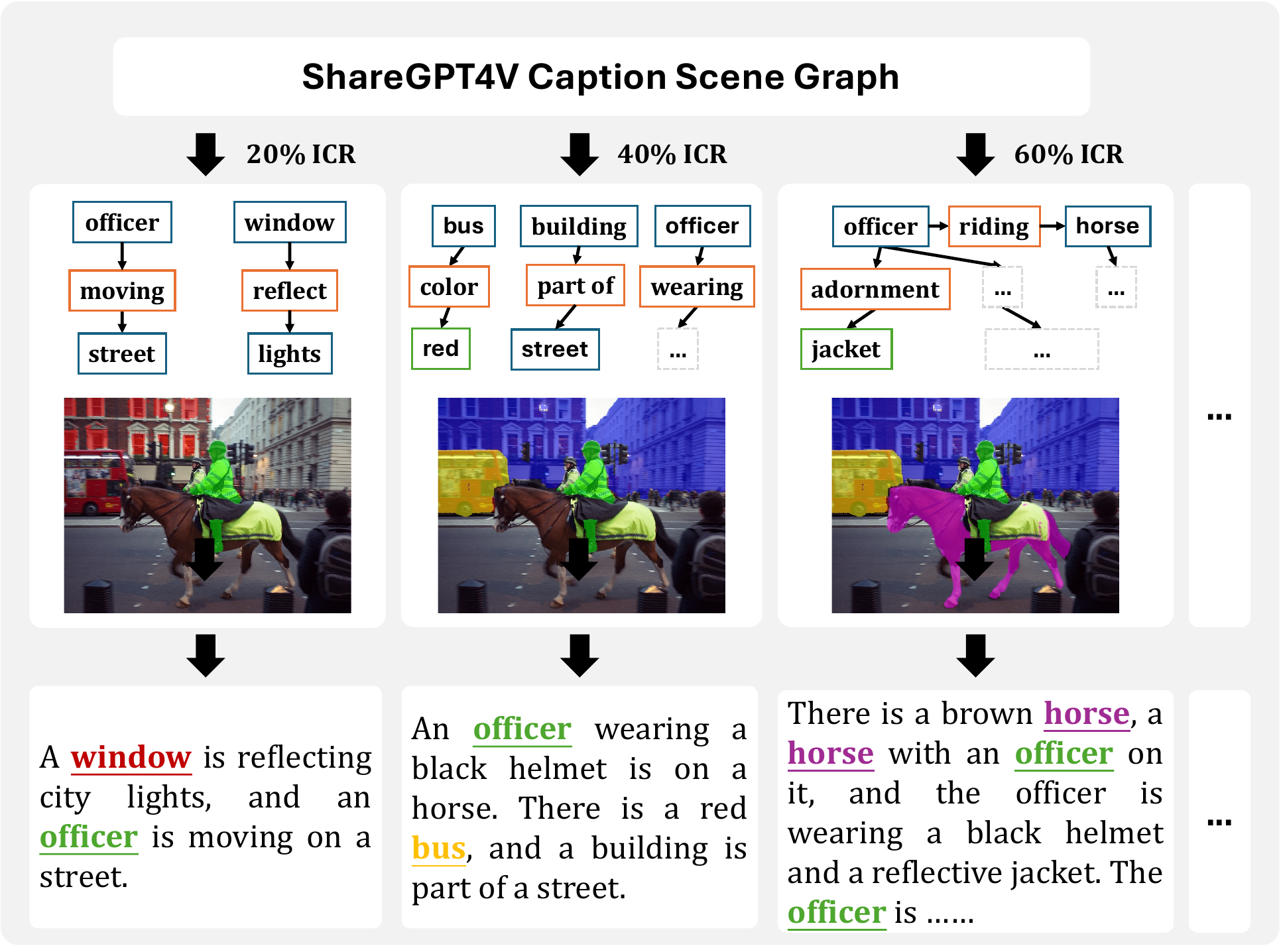}
    \caption{An illustration of how scene graphs are sampled to produce captions by varying image coverage rates (ICRs). Starting from the scene graph of a detailed caption. Then sub-graphs of different ICRs are sampled and converted into captions.}
    \label{fig:sample_sg_process}
\end{figure}

\begin{figure*}[ht!]
    \centering
    \includegraphics[width=\linewidth]{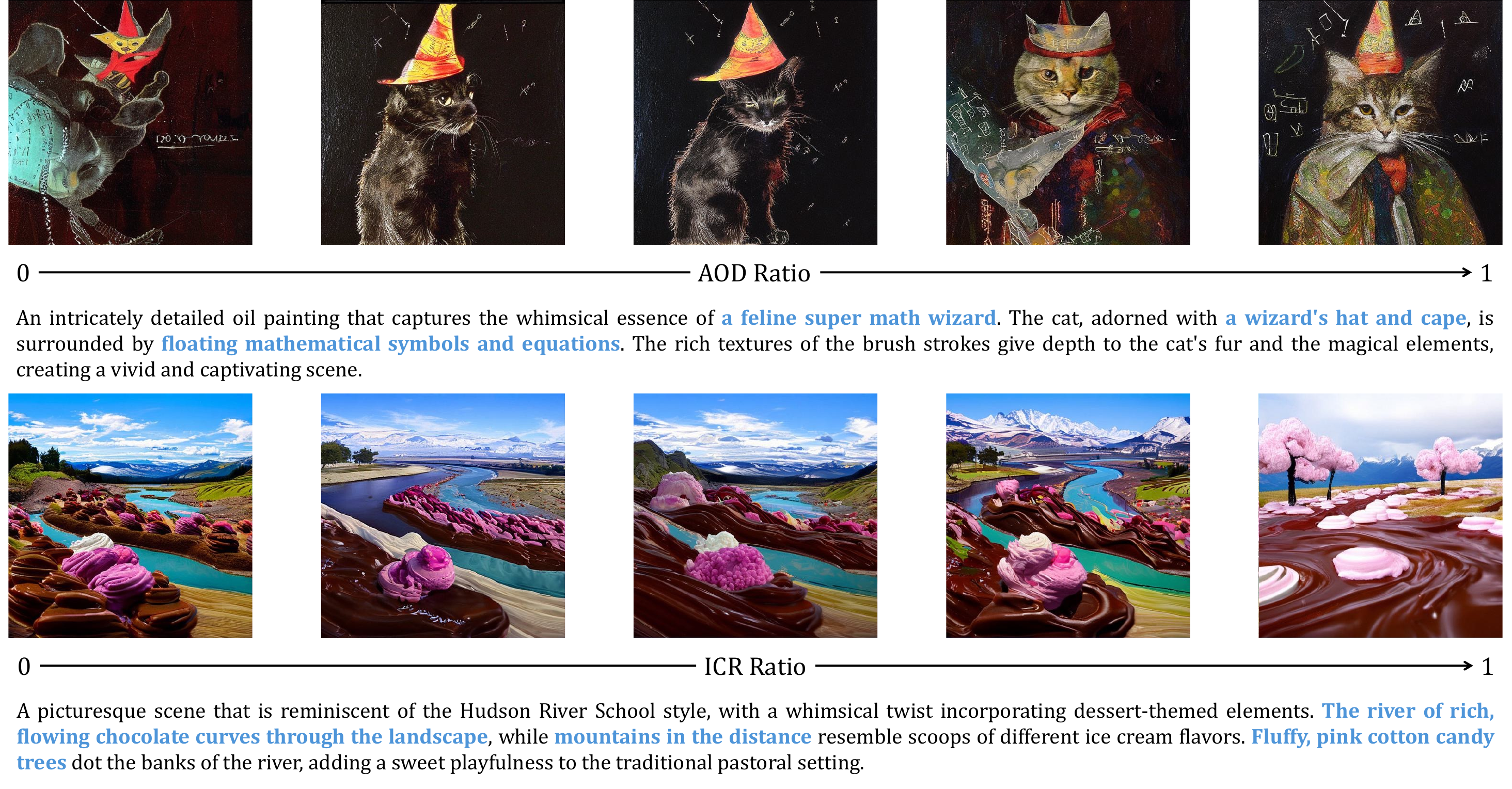}
    \caption{Generation examples of models trained by captions of different ICR ratios and AOD ratios using DPG prompts.}
    \label{fig:gen_example_1}
\end{figure*}

\subsection{Base Model and Datasets}

We select \textbf{Lumina-Next-T2I} \cite{zhuo_lumina-next_2024}, an open-source text-to-image (T2I) model, as our base model, which utilizes Next-DiT (a 2B-parameter diffusion transformer) as the backbone and Gemma-2B \cite{team_gemma:_2024} as the text prompt encoder (support long text prompt encoding). 
For comprehensive evaluation, we employ multiple datasets. The \textbf{MSCOCO} 2017 training subset \cite{fleet_microsoft_2014} is used with detailed captions from ShareGPT4V \cite{leonardis_sharegpt4v:_2025}. We partition it into 5,000 image-text pairs for testing while allocating the remaining samples for training to assess in-domain generation performance. For out-of-domain generalization evaluation, we incorporate the \textbf{Dense Prompt Graph} (DPG) benchmark, which assesses the consistency between generated visual content and input long prompts. Additionally, we utilize the \textbf{ImageInWords} (IIW) dataset \cite{garg_imageinwords:_2024}, which consists of 400 human-curated hyper-detailed image descriptions created through a rigorous human-in-the-loop annotation framework. It can facilitate precise evaluation of text-to-image reconstruction capabilities, particularly for intricate visual details.

\subsection{Metrics}
\label{sec:metrics}
We evaluate the text-to-image model based on three key aspects: (1) the overall quality and realism of the generated images; (2) the alignment between the generated image and the text prompt; (3) the model’s ability to reconstruct the original image from a detailed textual description. For image realism, we use the Fréchet Inception Distance (FID), where lower values indicate higher fidelity to real images. The CLIP Score (CLIP-S) measures semantic similarity between the generated image and the text, while CLIP Image Similarity (CLIP-IS) evaluates the resemblance between the generated and original reference images. Additionally, the Dense Prompt Graph (DPG) Benchmark employs a question-answering framework using the mPLUG-large model to quantify the alignment between the generated image and the input prompt. The detailed introduction for each metric is listed in the supplementary material.

\subsection{Influence of ICR and AOD}

\begin{figure}[t!]
    \centering
    \includegraphics[width=\linewidth]{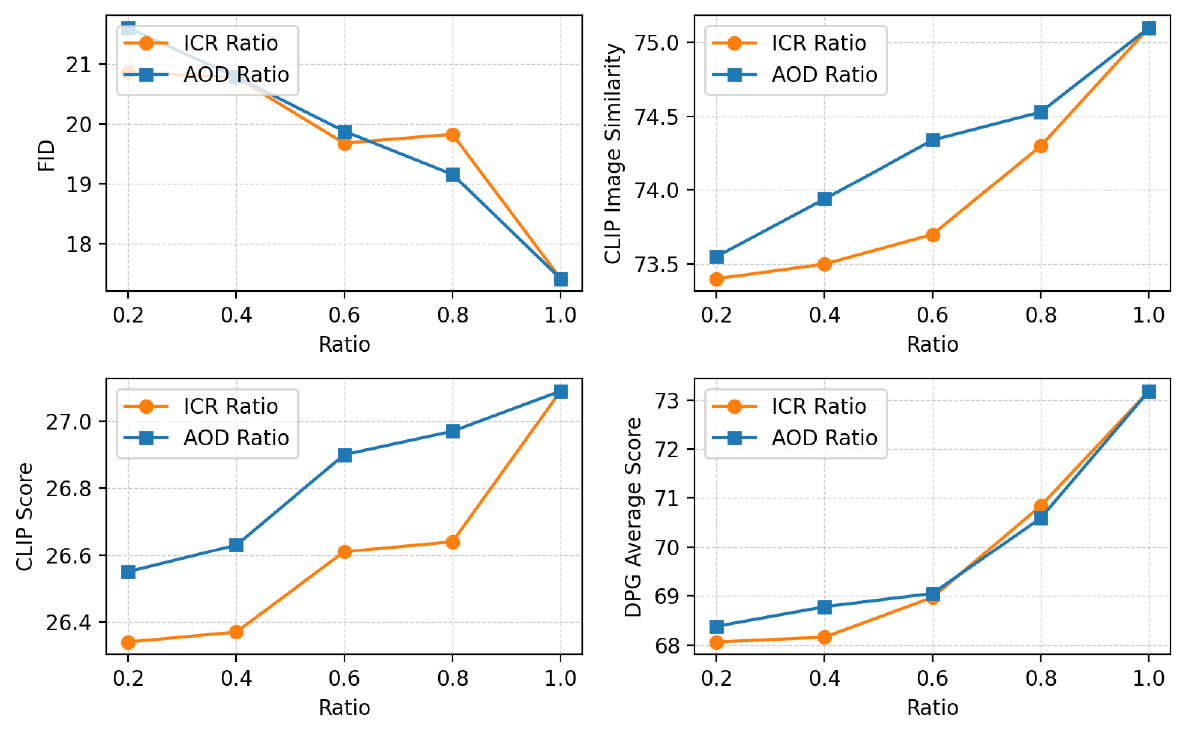}
    \caption{Text-to-image performance of Lumina-Next-T2I fine-tuned by different ratios of ICR and AOD captions.}
    \label{fig:pilot_line_plot}
\end{figure}

To investigate the individual effects of ICR and AOD, we generate captions with varying ICR and AOD values for the same image. We start with the original training set captions and parse them into scene graphs. From these graphs, we sample sub-graphs to create distinct captions with controlled variations in ICR and AOD as shown in \Cref{fig:sample_sg_process}.
Specifically, we iteratively sample sub-graphs corresponding to $20\% (\pm 5\%)$, $40\% (\pm 5\%)$, $60\% (\pm 5\%)$, and $80\%(\pm 5\%)$ of the original ICR and AOD. Then we synthesize these sub-graphs into captions. Increasing the ICR results in captions that cover a larger area of the image while increasing the AOD leads to captions that include more attributes for each object and more detailed relationships between objects. Finally, these synthetic captions are used to train Lumina-Next-T2I for performance comparison.

\vspace{-1em}

\paragraph{Positive Correlation Trend.} 
As shown in \Cref{fig:pilot_line_plot}, our experiments reveal that both ICR and AOD have a consistent positive correlation with the T2I performance. Specifically, increasing the ICR and AOD values in training captions led to progressively better results across various metrics. More detailed text conditions resulted in lower FID scores, indicating improved generation quality, while higher CLIP-IS values suggested greater similarity between generated and original images. Additionally, the rising CLIP-S and DPG scores confirmed enhanced image-text alignment. We observed that captions with a $60\%$ ICR sometimes outperformed those with $80\%$ ICR on certain metrics, such as CLIP-S and FID. This may be due to the omission of small but important objects in images when using the 80\% relative ICR setting. We further show some generation examples in \Cref{fig:gen_example_1}. The model trained by higher AOD and ICR usually has a relatively better generation quality and higher image-text consistency. With the increase of AOD, the details of each object are generated better, with the increase of ICR, the objects in the scene are more comprehensive, and the relationships between objects are also improved.


\begin{table}[t!]
    \footnotesize
    \tabcolsep=3.5pt
    \centering
    \begin{tabular}{llcccccc}
        \toprule
        & Ratio & Average & Attribute & Relation & Entity & Global & Other\\
        \midrule
         \multirow{5}{*}{ICR} & 0.2 & 68.06 & 78.66 & 80.36 & 79.78 & 79.27 & 80.57 \\
         & 0.4 & 68.16 & 79.80 & 80.02 & 79.70 & 77.26 & 80.40\\
         & 0.6 & 68.97 & 79.42 & 81.17 & 81.21 & 83.68 & 79.20 \\
         & 0.8 & 70.84 & 80.13 & 82.87 & 80.38 & 86.71 & 84.71\\
         & 1 & 73.18 & 83.07 & 85.75 & 83.11 & 84.85 & 82.08 \\
        \midrule

        \multirow{5}{*}{AOD} & 0.2 & 68.38 & 79.39 & 80.21 & 80.90 & 77.37 & 80.77 \\
        & 0.4 & 68.78 & 80.94 & 81.06 & 80.50 & 79.89 & 82.46 \\
        & 0.6 & 69.05 & 80.34 & 83.79 & 79.26 & 80.79 & 77.95 \\
        & 0.8 & 70.59 & 81.85 & 83.01 & 82.14 & 80.64 & 81.34 \\
        & 1 & 73.18 & 83.07 & 85.75 & 83.11 & 84.85 & 82.08 \\
        \bottomrule

    \end{tabular}
    \caption{DPG performance comparison under different ICR / AOD ratios where ratio 1 represents the ShareGPT4V captions.}
    \label{tab:icr-dpg-l1}
\end{table}

\begin{figure}[t!]
    \centering
    \includegraphics[width=0.9\linewidth]{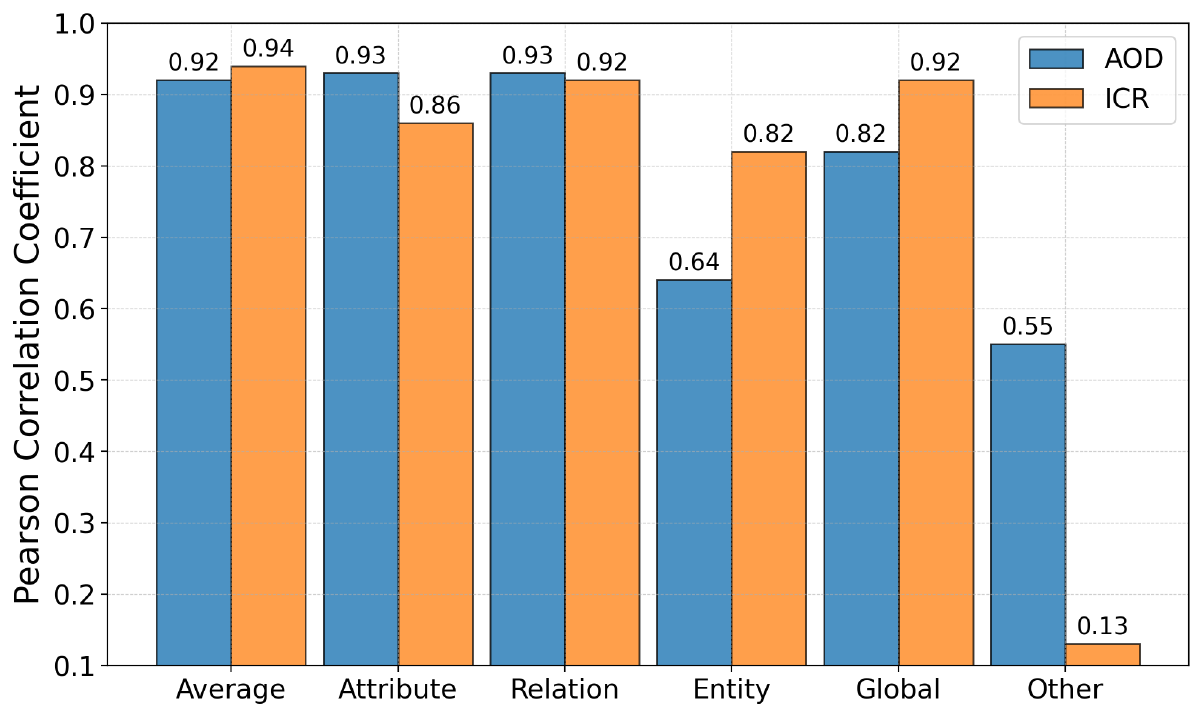}
    \caption{Comparison of Pearson correlation coefficients between ICR/AOD and DPG L1 categories' scores.}
    \label{fig:pearson}
\end{figure}


\vspace{-1em}

\paragraph{Influence on Different Dimensions.}

We present the scores of DPG L1 categories in \Cref{tab:icr-dpg-l1}. To analyze the influence of ICR and AOD on different dimensions, we compute the Pearson correlation coefficient $r$ between the AOD/ICR ratio and the scores of various dimensions. The results in \Cref{fig:pearson} indicate that ICR exhibits a higher Pearson correlation with entity and global generation performance compared to AOD. This suggests that the generation performance of these two dimensions is more strongly influenced by ICR, as captions with higher ICR contain more object information and more area. Conversely, attribute performance is more significantly impacted by AOD, as captions with higher AOD include more attributes per object. In terms of relations, both ICR and AOD show strong correlations. This is because an increase in ICR captures more objects, thereby capturing their interrelations. And more relations contribute to a higher AOD.



\begin{figure}
    \centering
    \includegraphics[width=\linewidth]{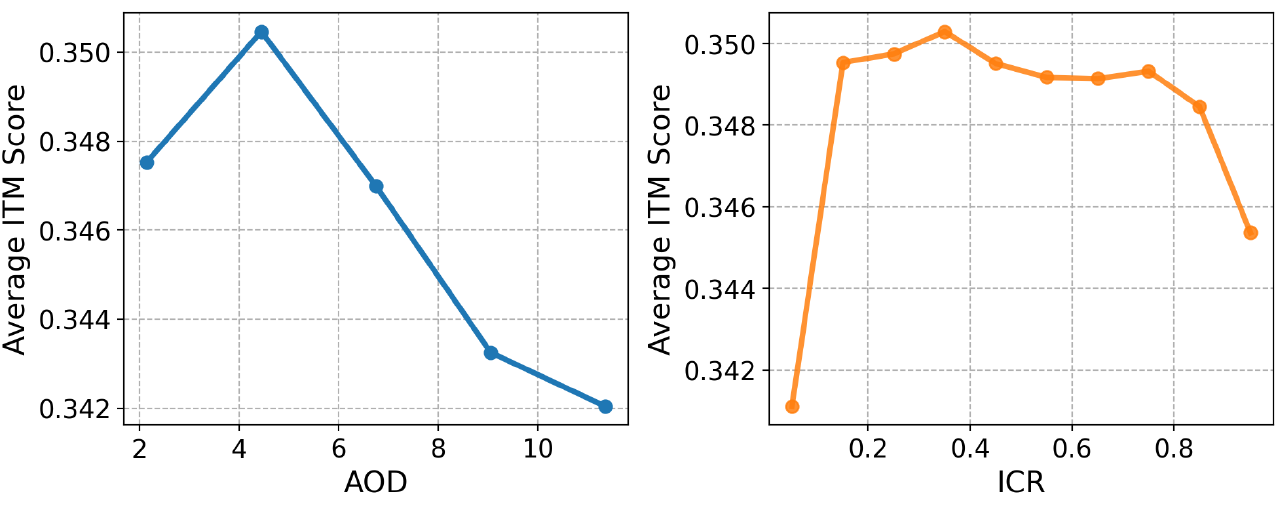}
    \caption{Impact of AOD and ICR on ITM Score: Captions with higher values of AOD and ICR may introduce more noise.}
    \label{fig:ITM-AOD-ICR}
\end{figure}

\begin{table*}[t]
\footnotesize
\begin{tabular}{lcccccccccccc}
\toprule
\multirow{2}{*}{Strategy} & \multicolumn{1}{c}{\multirow{2}{*}{num}} & \multicolumn{1}{c}{\multirow{2}{*}{avg. ICR}} & \multicolumn{1}{c}{\multirow{2}{*}{avg. AOD}} & \multicolumn{3}{c}{MSCOCO} & \multicolumn{6}{c}{DPG $\uparrow$}                 \\
& \multicolumn{1}{c}{} & \multicolumn{1}{c}{} & \multicolumn{1}{c}{} & FID $\downarrow$ & CLIP-S $\uparrow$ & CLIP-IS $\uparrow$ & Average & Attribute & Relation & Entity & Global & Other \\ \midrule

Full   & 113,162 & 76.75 & 2.19 & 17.42 & 27.09 & 75.10 & 73.18 & 83.07 & \textbf{85.75} & 83.11 & 84.85 & 82.08 \\ 

Rand & 20,000 & 76.74 & 2.18 & \textbf{15.97} & 27.43 & \textbf{75.74} & 74.78 & \underline{84.12} & 84.60 & 81.91 & 84.0 & 81.91 \\

Len & 20,000  & 73.37 & 2.08 & \underline{16.30} & 27.39 & 75.53 & 75.61 & 83.81 & \underline{85.55}  & 83.74 & \underline{87.16} & \textbf{85.39} \\ 

ITM-Len & 20,000 & 73.92  & 2.15 & 16.41 & \underline{27.52} & 75.53 & \underline{75.62} & \textbf{85.13} & 83.08 & \underline{84.86} & 77.80 & 84.21\\


\rowcolor{RoyalBlue!30} \textbf{Ours} & 20,000 & 84.07 & 2.36 & 16.32 & \textbf{27.63} & \underline{75.73} & \textbf{76.28} & 83.62 & 85.28 & \textbf{85.45} & \textbf{87.70} & \underline{84.99} \\

\bottomrule        
\end{tabular}
\centering
\caption{T2I performance of various data selection strategies. Best scores are highlighted in bold, and second-best scores are underlined.}
\label{tab:cap-filtering}
\end{table*}

\begin{table*}[ht]
\centering
\footnotesize
\begin{tabular}{cccccccccccccc}
\toprule
\multicolumn{3}{c}{Filtering Metrics} & \multirow{2}{*}{avg. ICR} & \multirow{2}{*}{avg. AOD} & \multicolumn{3}{c}{MSCOCO} & \multicolumn{6}{c}{DPG $\uparrow$}                 \\
ITM & ICR & AOD &  & & FID $\downarrow$ & CLIP-S $\uparrow$ & CLIP-IS $\uparrow$ & Average & Attribute & Relation & Entity & Global & Other \\ \midrule
\checkmark & & & 74.89 & 2.23 & 16.52 & 27.58 &  75.40 & \underline{75.98} & 84.31 & 84.31 & 84.11 & 86.50 & \textbf{86.34}\\ 

\checkmark & \checkmark &  & 88.15 & 2.19 & \underline{16.04} & 27.60 & 75.57 & 75.97 & \textbf{84.70} & \textbf{87.53} & 84.04 & 85.40 & 83.54  \\ 

\checkmark & & \checkmark & 74.30 & 2.51 & 16.22 & \underline{27.62} & \underline{75.65} & 75.41 & 84.25 & 86.47 & \underline{84.20} & 81.79 & 84.33 \\ 


& \checkmark & \checkmark & 92.70 & 2.92 & 16.32   & 27.38 & 75.49 & 74.52 & \underline{84.54} & \underline{87.43} & 83.36 & 82.23 & 81.76 \\ 

\checkmark & \checkmark & \checkmark & 85.54 & 2.36 & \textbf{15.68}  & 27.59 & 75.60 & 75.62 & 83.59 & 85.25 & 84.01  & \underline{87.11} & 81.84 \\ 
\midrule
\multicolumn{10}{l}{+ Length Normalization} \\
\rowcolor{RoyalBlue!30} \checkmark & \checkmark & \checkmark & 84.07 & 2.36 & 16.32 & \textbf{27.63} & \textbf{75.73} & \textbf{76.28} & 83.62 & 85.28 & \textbf{85.45} & \textbf{87.70} & \underline{84.99} \\

\bottomrule
\end{tabular}
\caption{Ablation results of metrics combinations. Best scores are highlighted in bold, and second-best scores are underlined.}
\label{tab:ablation}
\end{table*}

\begin{figure*}[ht]
     \centering
    \vspace{-1em}
    \includegraphics[width=0.95\linewidth]{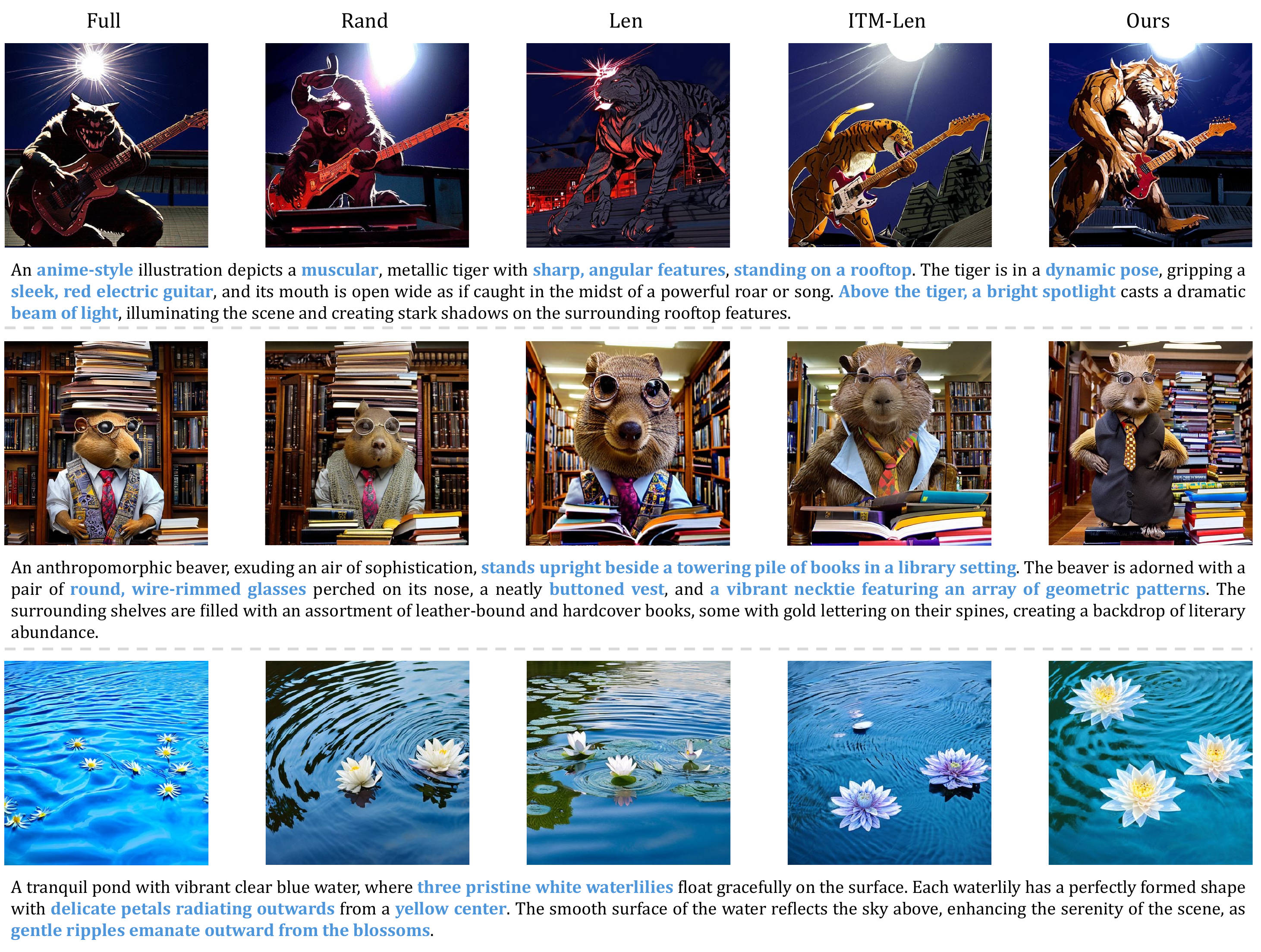}
    \caption{Generated image examples from DPG benchmark. The blue text highlights matching details generated by our method.}
    \vspace{-1.1em}
    \label{fig:vis-iiw}
\end{figure*}

\begin{table}[t]
    \vspace{-1em}
    \footnotesize
    \centering
    \begin{tabular}{lcccc>{\columncolor{RoyalBlue!30}}c}
    \toprule
    Metrics  & Full  & Rand  & Len   & ITM-Len & Ours \\
    \midrule
    CLIP-S   & 25.89 & 26.16 & 26.17 & \underline{26.22} & \textbf{26.36} \\
    CLIP-IS  & 67.20 & 68.27 & 68.04 & \underline{68.45} & \textbf{68.72} \\
    
    \bottomrule
    \end{tabular}
    \caption{Results on 400-IIW test set. Best scores are highlighted in bold, and second-best scores are underlined.}
    \label{tab:iiw-clipis}

\end{table}

\subsection{Training Caption Filtering}

The experimental design presented in the previous section primarily relies on well-engineered synthetic captions, which may not fully represent real-world training scenarios utilizing MLLM-generated descriptions. While the average ICR and AOD of ShareGPT4V have already reached 76.75\% and 2.19/object, respectively, an open question remains: can performance be further enhanced by using more detailed captions? Moreover, is full training data necessary, or can a smaller set of highly detailed caption-image pairs suffice? To address these questions and validate the practical effectiveness of ICR and AOD metrics in actual data curation, we conducted systematic caption filtering experiments on the ShareGPT4V dataset.

Notably, more important than detailness is the caption's semantic correctness, which constitutes the fundamental requirement for training data quality. Detailed but erroneous descriptions may induce amplified noise propagation due to their high information density. To quantify this trade-off between semantic correctness and detailness, we apply LLM2CLIP \cite{wu_llm2clip:_2024}, an enhanced vision-language model that replaces CLIP's text encoder with a large language model. This framework enables the calculation of image-text matching (ITM) scores on image-detailed caption pairs. We use this metric to quantify the semantic correctness.

Through comprehensive annotation of ICR, AOD, and ITM metrics across the training set captions, we performed analysis by grouping descriptions with comparable ICR/AOD levels. As shown in \Cref{fig:ITM-AOD-ICR}, the ITM scores exhibit an inverted-U relationship with both ICR and AOD values: initially improving with increasing metric values before reaching an optimal plateau and subsequently degrading at elevated levels. This non-monotonic relationship suggests that while moderate detailness enhances alignment quality, excessively detailed ShareGPT4V captions may contain perceptual inaccuracies or hallucinated content.

\paragraph{Our Caption Selection Strategy and Baselines.} Based on previous observations, we build our caption filtering strategy through sequential filtering to trade between detailness and semantic correctness: \textbf{(1) Semantic pruning}: First, we filter the dataset by retaining only the $K$ samples with the highest ITM scores, thereby eliminating misaligned pairs. \textbf{(2) Detailness ranking}: Using caption detailness formulated in \Cref{eq:detailness} for the filtered subset, selecting the top $T$ samples to maximize descriptive richness while preserving semantic fidelity.  
We compare our approach against the following data selection strategies: (1) \textbf{Full data}: Training on the entire unfiltered dataset. (2) \textbf{Random}: Randomly selecting $T$ samples from the full dataset. (3) \textbf{Length}: Selecting $T$ samples with the longest captions by word count. (4) \textbf{ITM + Length}: Selecting $T$ longest samples from the $K$ samples with the highest ITM scores. This baseline is used to compare the role of detailness with pure caption length. 

Here we set $K$ as 30,000 and $T$ as 20,000. We compare the average ICR and average AOD of training data from different data selection strategies in \Cref{tab:cap-filtering}. Notably, compared to data selected using length-based strategies, the subset chosen by our method exhibits higher ICR and AOD. This suggests that longer captions do not necessarily provide high image coverage or richer semantic representations of attributes and relationships.

\vspace{-1em}

\paragraph{Results.}

The results of different caption selection strategies are presented in \Cref{tab:cap-filtering}. Surprisingly, random selection outperforms using the full dataset, suggesting that the full data captions may contain substantial noise. Compared to selecting solely based on length, choosing the longest caption among those with the highest ITM score yields a higher CLIP-S. Furthermore, our method surpasses ITM-Len in multiple metrics, achieving a higher DPG average score, CLIP-IS, and CLIP-S while maintaining a lower FID. We visualize several generated examples in \Cref{fig:vis-iiw}. Images generated using our data selection method exhibit better image-text alignment and capture fine details more accurately. For instance, in the second example, the prompt specifies a buttoned vest, yet beavers generated by other selection methods wear vests without buttons. Similarly, in the third example, our method successfully adheres to the prompt’s specification regarding the number of water lilies. We provide more generated examples in supplementary materials. Additionally, we evaluate different data selection strategies on the 400-image IIW test set, an out-of-domain dataset, with results shown in \Cref{tab:iiw-clipis}. Our method achieves the highest CLIP-S and CLIP-IS scores, demonstrating that models trained with our filtering strategy are better able to reconstruct the fine details of the out-of-domain images using text prompts alone and the generated image is more consistent with the input text.

\vspace{-1em}

\paragraph{Ablations.}

We further conduct ablation studies on the filtering metrics used in our selection strategy (see \Cref{tab:ablation}). First, we evaluate a variant that relies exclusively on the ITM score. While this ITM-only approach yields a relatively high DPG average score and CLIP-S—indicating improved image-text consistency—it also results in a higher FID and lower CLIP-IS. These findings suggest that relying solely on ITM is insufficient for achieving high-quality image reconstruction. When combining ITM with AOD, the DPG score worsened, implying that captions with high AOD introduce noise and adversely affect image-text consistency. In contrast, the ITM + ICR strategy outperforms ITM + AOD, as captions with higher ICR introduce less noise, consistent with the observations in \Cref{fig:ITM-AOD-ICR}. As expected, filtering solely with ICR and AOD leads to even poorer performance, underscoring the need for a combined metric approach. Finally, normalizing detailness by caption length further enhances performance, indicating that ShareGPT4V's COCO image descriptions contain substantial redundant non-visual information that hinders effective image-text alignment during model training.

\subsection{Training Captions Annotation Suggestions}

Based on our experimental findings, we recommend the following guidelines for image caption annotation: (1) Avoid using caption length as the sole measure of detail. Instead, evaluate the annotation by considering how comprehensively it covers the image and the specificity of the descriptions for each object. (2) Prioritize annotations accuracy. Ensure descriptions are precise and reliable before adding detail, as overly detailed yet inaccurate captions may introduce noise and hinder image-text alignment.

\section{Conclusion}

In this work, we proposed a new metric considering two aspects—image coverage rate (ICR) and average object detailness (AOD)—to re-evaluate caption detailness from a new perspective, moving beyond the traditional reliance on caption length. Our experiments on the COCO dataset with ShareGPT4V captions demonstrate that T2I models trained on captions with higher ICR and AOD yield significantly improved image-text consistency and reconstruction quality. Moreover, filtering training data using these metrics outperforms both full-dataset and length-based filtering approaches, achieving superior results with no more than 20\% of the full data. These findings underscore the importance of using nuanced, detail-aware metrics for annotating and selecting captions in text-to-image generation.

{
    \small
    \bibliographystyle{ieeenat_fullname}
    \bibliography{main}
}

\end{document}